\documentclass[conference]{IEEEtran}
\usepackage{times}

\usepackage[numbers]{natbib}
\usepackage{multicol}
\usepackage[bookmarks=true]{hyperref}

\usepackage{graphicx}
\usepackage{color}

\usepackage{amsmath}
\usepackage{amssymb}
\usepackage{bm}
\usepackage{subcaption}
\usepackage{multirow}

\usepackage{rss2026ishigaki}

\usepackage[final]{changes}

\begin{document}

\title{Implicit Null-space Manifold Generation for Redundant Robotic Systems}


\author{
Taiki Ishigaki\\
Tokyo University of Science, Japan\\
\texttt{taiki.ishigaki@rs.tus.ac.jp}
\and
Teresa Vidal-Calleja\\
University of Technology Sydney, Australia
\and
Ko Ayusawa \\
National Institute of Advanced Industrial Science and Technology, Japan
\and
Eiichi Yoshida \\
Tokyo University of Science, Japan
}


%

\maketitle

\begin{abstract}
Robotic systems with redundant degrees of freedom can achieve the same task outcome using multiple configurations, resulting in solution sets that form manifolds in the configuration space.
Existing approaches typically exploit such redundancy locally through Jacobian-based techniques to compute individual solutions or trajectories. While effective for solution computation, these methods do not retain a representation of the geometry of the solution set itself.
In this work, we adopt a representation-centric approach to estimate the geometric structure of the solution space.
We consider solution manifolds induced by general task-defining maps and
construct an implicit scalar field over the configuration space, whose
zero-level set corresponds to the solution manifold.
To this end, we generate samples in the neighborhood of the solution manifold using a Jacobian-guided exploration strategy, which efficiently captures its local and global structure.
The resulting implicit representation is defined over the configuration space and naturally induces a continuous, distance field that encodes proximity to the solution manifold. Experiments on a planar three-link robot and a seven-degree-of-freedom Franka manipulator demonstrate the effectiveness of the proposed representation.
Furthermore, the framework enables consistent modeling of solution spaces across
families of tasks with continuous variation.
\end{abstract}

\IEEEpeerreviewmaketitle

\section{Introduction}
Robotic systems often exhibit redundancy, where multiple configurations yield the same task outcome.
Such redundancy naturally induces solution sets in the configuration space and plays a fundamental role in inverse kinematics~\cite{nakamura1986inverse, sugihara2011solvability}, redundancy resolution, and motion generation.
These problems are typically addressed in a query-driven manner, where feasible configurations or trajectories are computed for individual task instances.

Existing methods, in both kinematic \cite{nakamura1990advanced} and dynamic \cite{khatib2003unified} settings, primarily exploit redundancy via local Jacobian-based techniques to compute solutions or motions.
While effective for obtaining feasible instances, these approaches do not provide an explicit representation of the underlying solution space.
As a result, the geometric structure of the null-space is discarded once a solution is obtained, limiting reuse across tasks and subsequent queries.
Our approach is motivated by classical results on self-motion manifolds \cite{burdick1989inverse, luck1993redundant}, which establish that redundancy-induced solution sets form smooth manifolds in the configuration space.

In this work, we adopt a representation-centric perspective on redundancy, focusing on modeling the geometric structure of solution spaces rather than computing individual solutions.
We consider tasks defined by a differentiable mapping
\begin{align}
    \bm{p} = \bm{f}(\bm{q}),
\end{align}
where $\bm{q} \in \mathbb{R}^{n}$ denotes configuration variables and $\bm{p} \in \mathbb{R}^{m}$ denotes task variables.
Importantly, the mapping $\bm{f}$ is not restricted to forward kinematics; it may represent any task-defining function for which function and Jacobian evaluations are available.

\begin{figure}[t]
    \centering
    \includegraphics[width=1.0\linewidth]{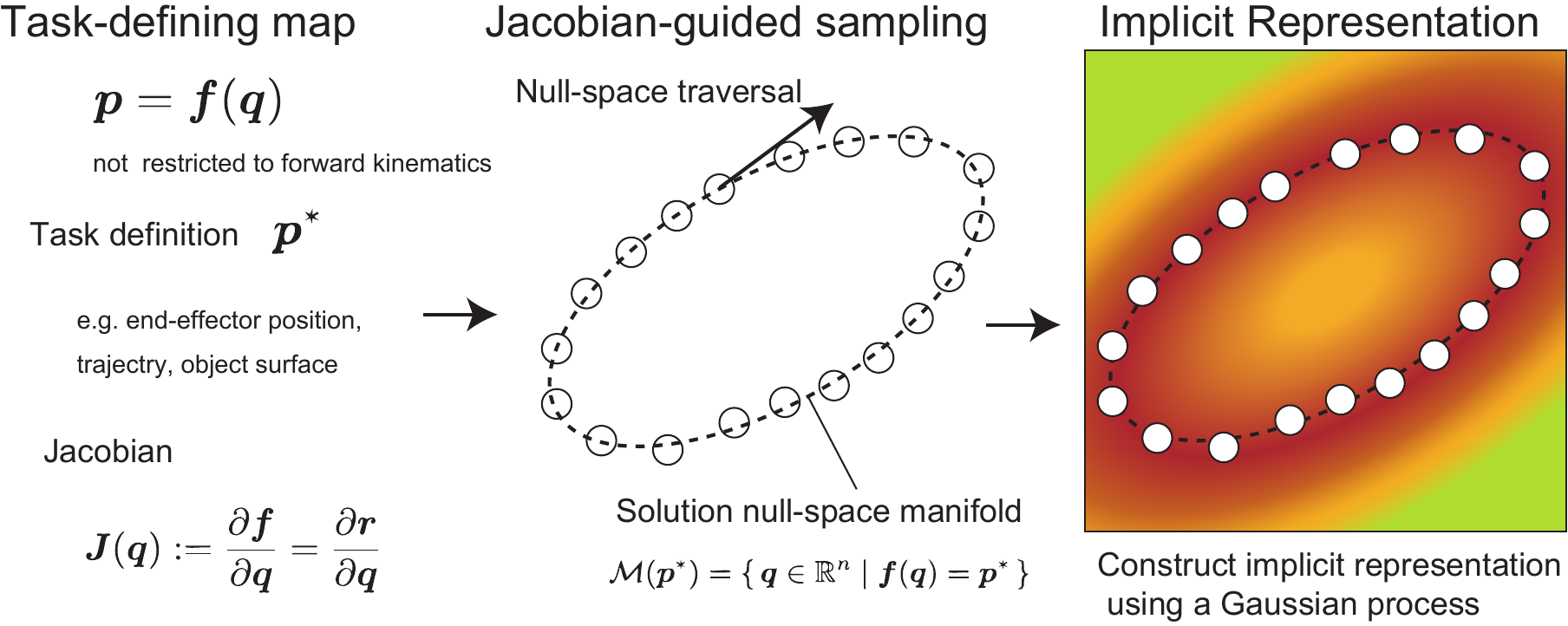}
    \caption{Overview of the proposed framework.
From a task-defining map $p=f(q)$, Jacobian-guided sampling produces samples in the neighborhood of the solution manifold in configuration space. An implicit representation is then constructed using a Gaussian process, providing a configuration-space model whose zero-level set represents the solution manifold.}
    \label{fig:overview}
\end{figure}

For a given target task value $\bm{p}^{\ast}$, the set of configurations satisfying $\bm{f}(\bm{q}) = \bm{p}^{\ast}$ forms an implicitly defined solution manifold,
\begin{align}
    \mathcal{M}(\bm{p}^{\ast})
    = \{\, \bm{q} \in \mathbb{R}^{n} \mid \bm{f}(\bm{q}) = \bm{p}^{\ast} \,\}.
\end{align}
It has been shown that, under redundancy, inverse kinematic solution sets form smooth self-motion manifolds in the configuration space \cite{burdick1989inverse}.
Recent progress has also been made toward directly computing such manifolds as global solution traces, for example via ODE-based integration in the Jacobian null space together with mechanisms for handling sign ambiguities and searching disconnected components \cite{guri2025ode}.
However, trace-based approaches primarily yield instance-specific point sets and do not produce a reusable configuration-space representation that supports geometric queries or downstream planning.

We propose a representation-centric framework for modeling these solution manifolds.
Specifically, we construct a probabilistic implicit representation whose zero-level set corresponds to $\mathcal{M}(\bm{p}^{\ast})$.
The implicit function $\bm{\phi}(\bm{q})$ is modeled as a Gaussian process (GP) and is constructed from configuration samples generated on and near the manifold.
Sampling is guided by the Jacobian of the task-defining map, which encodes local null-space geometry, enabling informative sampling without exhaustive exploration of the configuration space or distance supervision.
Once constructed, the resulting implicit representation can be reused for a fixed task instance to support proximity evaluation and motion generation, and it induces a continuous, distance field over the configuration space that reflects proximity to the solution manifold.

Although inverse kinematics provides a convenient testbed in our experiments, the proposed framework does not aim to solve inverse kinematics problems.
Instead, it targets reusable representation of solution spaces induced by general task-defining maps, enabling proximity queries and task-aware sampling beyond inverse kinematics.

\section{Related Work}

\subsection{Inverse Kinematics and Redundancy Resolution}


Inverse kinematics (IK) \cite{nakamura1986inverse, sugihara2011solvability} and redundancy resolution have been extensively studied as means of computing configurations or trajectories that satisfy task constraints.
Classical approaches exploit the Jacobian null-space to locally resolve redundancy and generate feasible motions.
These methods are inherently query-driven, focusing on computing feasible solutions for individual task instances.


\subsection{Self-Motion Manifolds and Geometric Characterization}




The geometric structure of solution sets induced by redundancy has been studied from differential geometric and topological perspectives.
Burdick \cite{burdick1989inverse} characterized inverse kinematic solution sets of redundant manipulators as self-motion manifolds, analyzing their dimension and topology.
More recently, self-motion manifolds have been explicitly computed as global solution traces via ODE-based integration of Jacobian null-space directions \cite{guri2025ode}.
These approaches provide accurate manifold traces for specific task instances.

\subsection{Learning-Based Inverse Models}


Learning-based approaches have been proposed to approximate inverse mappings from task space to configuration space and to address multi-solution IK problems \cite{ames2022ikflow}.
These methods typically rely on dense training data and focus on predicting configurations directly.
Moreover, they do not explicitly exploit Jacobian-based geometric structure.

\subsection{Implicit Representations and Distance Fields}




Implicit surface representations and distance fields provide a framework for encoding geometry as scalar-valued functions over a space.
Gaussian process implicit surfaces (GPIS) have been used to model shapes~\cite{williams2006gaussian, wu2020skeleton, martens2016geometric}, and distance fields~\cite{legentil2024gpdf} without explicit parametrization.
More recently, configuration space distance fields (CDFs) have been proposed to encode task-relevant geometry directly in joint space \cite{li2024configuration}.
These representations are primarily designed for distance queries and motion planning.

\subsection{Position of This Work}


In contrast to the above approaches, this work focuses on representing the geometry of solution manifolds induced by general task-defining maps.
Rather than computing individual inverse solutions or learning direct inverse mappings, we construct a probabilistic implicit representation of the solution space itself.
By exploiting Jacobian-guided local geometric information, the proposed method bridges classical self-motion manifold theory and implicit representations, enabling reusable modeling of null-space geometry beyond inverse kinematics.

\section{METHOD}

\begin{figure}
    \centering
    \includegraphics[width=0.7\linewidth]{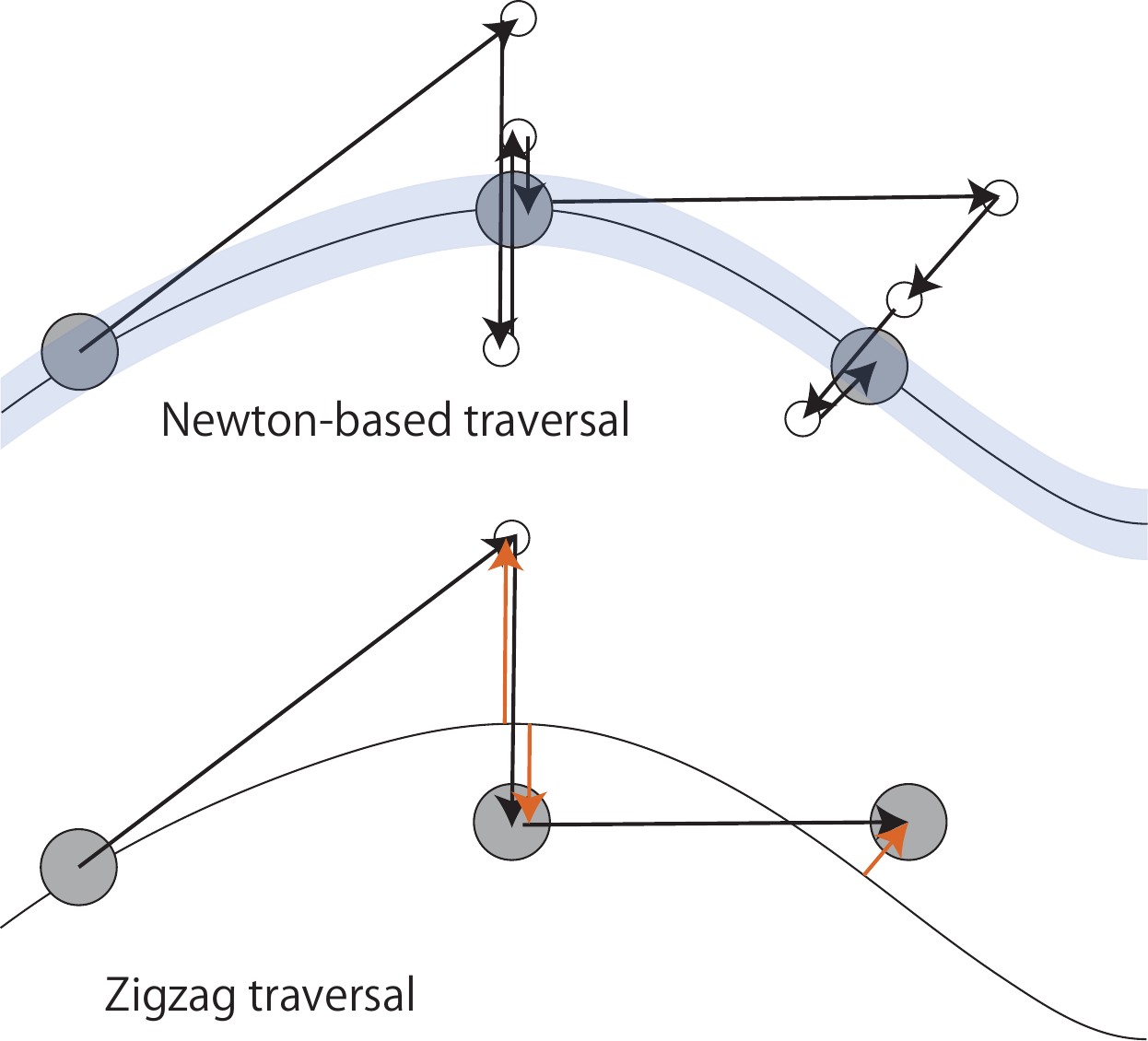}
    \caption{Jacobian-guided sampling for traversing a solution manifold.}
    \label{fig:jacobian_sampling}
\end{figure}

Given a differentiable task-defining map $\bm{p}=\bm{f}(\bm{q})$ with
$\bm{q}\in\mathbb{R}^n$ and $\bm{p}\in\mathbb{R}^m$, we fix a target $\bm{p}^\ast$
and define the residual
\begin{align}
\bm{r}(\bm{q}) := \bm{f}(\bm{q}) - \bm{p}^{\ast}.
\end{align}
The solution manifold is
\begin{align}
\bm{M}(\bm{p}^{\ast}) =
\{\bm{q}\in\mathbb{R}^n \mid \bm{r}(\bm{q})=\bm{0}\}.
\end{align}
\added{
Although $\bm{p}$ may lie in SE(3), we represent it as $\bm{p} \in \mathbb{R}^p$ to explicitly encode the dimensionality of the task, while the task error is computed in the corresponding tangent space (e.g., via $\log(H^{-1}H')$ for SE(3)).}
We assume access only to $\bm{r}(\bm{q})$ and its Jacobian
\begin{align}
\bm{J}(\bm{q}) := \frac{\partial \bm{f}}{\partial \bm{q}}
= \frac{\partial \bm{r}}{\partial \bm{q}}.
\end{align}
Since $\Delta\bm{p}=\bm{J}\Delta\bm{q}$, displacements in $\mathrm{Null}(\bm{J})$
do not affect the task to first order and thus span the local tangent space of
$\bm{M}(\bm{p}^{\ast})$.

Our framework has two components: (i) Jacobian-guided sampling of points on/near
$\bm{M}(\bm{p}^{\ast})$, and (ii) GP-based implicit modeling to obtain a
distance-like field in configuration space.

\subsection{Jacobian-Guided Sampling}
\label{jacob-guide-smaple}

Fig.~\ref{fig:jacobian_sampling} illustrates the proposed Jacobian-guided sampling strategy for traversing the solution manifold $\bm{M}(\bm{p}^{\ast})$ in configuration space.
Starting from a configuration that satisfies the task constraint, sampling proceeds by alternately (1) advancing along directions tangent to the manifold and (2) restoring task consistency through local correction.
The tangent directions are obtained from the null space of the Jacobian, which locally approximates the manifold geometry.
Two variants of this traversal are shown in Fig.~\ref{fig:jacobian_sampling}: a Newton-based method that tightly follows the manifold via frequent projection, and a zigzag method that relaxes projection accuracy and alternates across the manifold to accelerate exploration.
Since joint angle limits can be straightforwardly represented in the configuration space, we consider a configuration space in which each joint angle is constrained to lie within the interval $[-\pi,\pi]$.

\subsubsection{Initial Point Search}
We first obtain an initial configuration $\bm{q}_0$ satisfying
$\bm{r}(\bm{q}_0)=\bm{0}$ by iterating a Gauss--Newton update:
\begin{align}
\Delta \bm{q}
=
\underset{\Delta \bm{q}}{\operatorname{argmin}}
\left\|
\bm{r}(\bm{q})
+
\bm{J}(\bm{q}) \Delta \bm{q}
\right\|^2, \ 
\bm{q} \leftarrow \bm{q} + \Delta \bm{q},
\end{align}
until $\|\bm{r}(\bm{q})\|\le \varepsilon$.

\subsubsection{Jacobian-Guided Manifold Traversal}
\label{jacob_guide}
Given $\bm{q}_k$, we advance along a null-space direction and then restore
constraint satisfaction:
\begin{align}
\label{tangent_update}
\tilde{\bm{q}}_{k+1}
&=
\bm{q}_k + \beta \bm{v}_k,
\quad
\bm{v}_k \in \mathrm{Null}(\bm{J}(\bm{q}_k)), \\
\bm{q}_{k+1} &= \Pi(\tilde{\bm{q}}_{k+1}),
\end{align}
where $\Pi(\cdot)$ denotes an operator that maps a configuration to a neighborhood of the manifold.
The vector $\bm{v}_k$ is a nontrivial solution satisfying $\bm{J}(\bm{q}_k)\bm{v}_k = \bm{0}$.
In the following, we introduce two different realizations of this operator.
The parameter $\beta$ represents the sampling interval: larger values result in coarser sampling. However, for manifolds with strong nonlinearity, excessively large $\beta$ may cause the tracing process to deviate from the manifold.

Joint angles are wrapped to the interval $[-\pi, \pi]$ to account for angular periodicity during exploration.
quation~\eqref{tangent_update} is specialized for one-dimensional manifolds, where a single tangent direction is obtained and the termination condition for manifold exploration can be defined in a straightforward manner.
Accordingly, in our implementation, the exploration on a given manifold is terminated when a previously sampled point is found within a $\beta$-neighborhood of the current sampling point.
While this work focuses on the case of a single redundant degree of freedom, tangent directions of a manifold can in general be obtained from the null space of the Jacobian.
By developing appropriate exploration strategies and termination conditions, the proposed framework can be extended to exploration on higher-dimensional manifolds.

\subsubsection{Newton-Based Traversal}
In the Newton-based variant, $\bm{v}_k$ is recomputed at each step from the
current Jacobian, yielding stable manifold-following at the cost of more
frequent direction/projection updates.

\subsubsection{Zigzag Traversal}
To reduce projection effort, we propose a zigzag strategy that relaxes strict
projection and instead maintains successive samples that approximately lie on
opposite sides of the manifold.

We define a correction direction toward the manifold by the minimum-norm
linearized solution of $\bm{J}(\bm{q})\Delta\bm{q}=-\bm{r}(\bm{q})$:
\begin{align}
\label{zigzag_d}
\bm{d}(\bm{q}) = -\bm{J}(\bm{q})^{\sharp} \bm{r}(\bm{q}),
\end{align}
where $(\cdot)^\sharp$ is the pseudo-inverse. Starting from
$\tilde{\bm{q}}_{k+1}$, we perform a small number of correction steps
\begin{align}
\label{zigzag_proj_step}
\bm{q}_{\ell} = \bm{q}_{\ell-1} + \gamma\, \bm{d}(\bm{q}_{\ell-1}),
\end{align}
and terminate when the direction to the manifold flips:
\begin{align}
\label{inner_condition}
\bm{d}(\bm{q}_{\ell-1}) \cdot \bm{d}(\bm{q}_{\ell}) < 0.
\end{align}
Equation \eqref{inner_condition} checks the relative orientation of the
correction directions toward the manifold at the two consecutive iterations.
A negative inner product means that the angle between these vectors is obtuse,
i.e., the two directions point to opposite sides.
This is a simple indicator that the manifold has been crossed and lies between
$\bm{q}_{\ell-1}$ and $\bm{q}_{\ell}$ (up to the local linearization used to
compute $\bm{d}$), so we stop without enforcing exact convergence.
We set $\gamma>1$ in \eqref{zigzag_proj_step} to intentionally overshoot and
promote such crossings.

\subsection{Probabilistic Implicit Function Representation}

GP regression is a nonparametric method that has been widely used
for nonlinear regression problems.
It enables probabilistic estimation of unknown function values at arbitrary
query points from a limited number of samples that may include noise and
variability.

GPIS methods
\cite{williams2006gaussian, wu2020skeleton, martens2016geometric}
employ GP to estimate an implicit surface from a set of noisy
sampling points.
Using GPIS, it is also possible to construct a distance field from the implicit
surface for arbitrary query points \cite{legentil2024gpdf}. 

In this study, we utilize manifold neighborhood samples generated by the method
described in Section~\ref{jacob-guide-smaple} and represent, using a shifted GPIS, a field
that simultaneously encodes the manifold structure and a customizable notion of distance from the manifold. \added{Unlike the standard GPIS method, which uses a constant zero-mean function, we shift the mean to one. This field can be interpreted as a smooth occupancy field, or equivalently, as an implicit surface on the 1-level set. While the surface can be modelled directly using this occupancy field,~\cite{legentil2024gpdf}
proposed applying an inverse mapping to recover a distance field far away from the surface. This can be understood as reversing the effect of the kernel function, which essentially represents the distance between two points when only one surface point and one query point are involved.}

Let us consider a GP with one mean:
\begin{align}
\bm{\phi}(\bm{q}) \sim \mathcal{GP}(1, k(\bm{q},\bm{q}')),
\end{align}
where $k(\bm{q},\bm{q}')$ is a covariance function.
As the covariance function, we employ the squared exponential (SE) kernel
\begin{align}
k(\bm{q},\bm{q}')
=
\exp
\left(
-\frac{\lVert \bm{q} - \bm{q}' \rVert^2}{2\ell^2}
\right).
\end{align}
where $\ell$ is the length-scale. 
Assuming observation noise, by using the observation value  $\qob$, the covariance matrix is defined as
\begin{align}
\bm{K}
=
k(\qob,\qob) + \sigma^2 \eyem.
\end{align}

Function values at the sampling points are set to $\bm{y}=\bm{1}$.
The coefficient vector $\bm{\alpha}$ in the posterior mean, requirement to construct the GP-based implicit representation, is computed as
\begin{align}
\bm{\alpha} = \bm{K}^{-1} \bm{y}.
\end{align}
Using $\bm{\alpha}$, the GPIS value at an arbitrary query point $\bm{q}$ is
evaluated as
\begin{align}
\label{gp_inference}
\bm{\phi}(\bm{q})
=
\bm{k}(\qob, \bm{q})^{\mathrm{T}} \bm{\alpha}.
\end{align}

Since the GP is constructed from samples on the null-space
manifold, we introduce a threshold $\beta \sim 1$ and regard configurations
$\bm{q}$ satisfying $\bm{\phi}(\bm{q}) > \beta$ as belonging to the
null-space manifold.

The gradient of $\bm{\phi}(\bm{q})$ can be computed analytically as
\begin{align}
\label{map_gradient}
\nabla \bm{\phi}(\bm{q})
=
\sum_i
-
\alpha_i
k(\qob_i,\bm{q})
\frac{(\qob_i - \bm{q})}{\ell^2}
\end{align}
where
$\bm{\alpha} = [\alpha_1,\dots,\alpha_N]^{\trans}$ and
$\qob = [\qob_1,\dots,\qob_N]^{\trans}$.
Finally, following the kernel inversion in~\cite{legentil2024gpdf}, we can estimate a pseudo-Euclidean distance field in configuration space induced by the implicit function as
\begin{align}
\label{distance_form_manifold}
d(\bm{q})
=
\sqrt{
\max\ \left(
0,
-2 \ell^2 \log \bm{\phi}(\bm{q})
\right)
}.
\end{align}

\section{EXPERIMENTS}

\subsection{Comparison of Jacobian-guided Sampling Methods}

\begin{table}[t]
    \centering
    \caption{Comparison of manifold sampling methods}
    \scriptsize{
    \begin{tabular}{cl|ccccc}
        \multirow{3}{*}{\textbf{Method}} & \multirow{3}{*}{$\beta$}
         & \textbf{Sampling} & \textbf{Coverage}
         & \textbf{Mean of Norm}
         & \textbf{Sampling} \\
         && \textbf{Time} & \textbf{Volume} 
         & \textbf{of Residual} & \textbf{Size}\\
         && [ms] & $V$[$\text{m}^3$] 
         &[m] & $n [-]$\\
        \hline 
        \multirow{4}{*}{Newton-based} 
        & 1.5 &
        1.3 & 0.0056 
        & $4.6 \times 10^{-5}$ &11 \\
        & 1.0 &
        1.4 & 0.0076 
        & $2.4 \times 10^{-5}$ &15 \\
        & 0.5 &
        2.1 & 0.0148 
        & $1.7 \times 10^{-5}$ & 29 \\
        & 0.1 & 
        7.7 & 0.0734 
        &$4.0 \times 10^{-6} $ & 140 \\ 
         \hline
        \multirow{4}{*}{Zigzag} 
        & 1.5 & 
        1.1 & 0.0052 
        & $9.9 \times 10^{-4}$ & 11\\
        & $1.0$ & 
        1.3 & 0.0079 
        & $8.0 \times 10^{-4}$ & 15\\
        & 0.5 & 
        2.3 & 0.0146 
        & $1.4 \times 10^{-4}$ & 29\\
        & $0.1$ & 10.1 & 0.0735 
        & $4.0 \times 10^{-6} $ & 140\\
        \hline
        \multirow{4}{*}{Random IK} 
        & & 3.2 & 0.0080 
        & $1.0 \times 10^{-6}$ & 15 \\
        &  & 6.5 & 0.0143 
        & $1.0 \times 10^{-6}$ & 30 \\
        &  & 31.3 & 0.0525
        & $1.0 \times 10^{-6}$ & 150 \\ 
        & & 63.3 & 0.0751 
        &  $1.0 \times 10^{-6} $ & 300\\ 
    \end{tabular}
    }
    \label{tab:sampling}
\end{table}

\begin{figure}
    \centering
    \includegraphics[width=\linewidth]{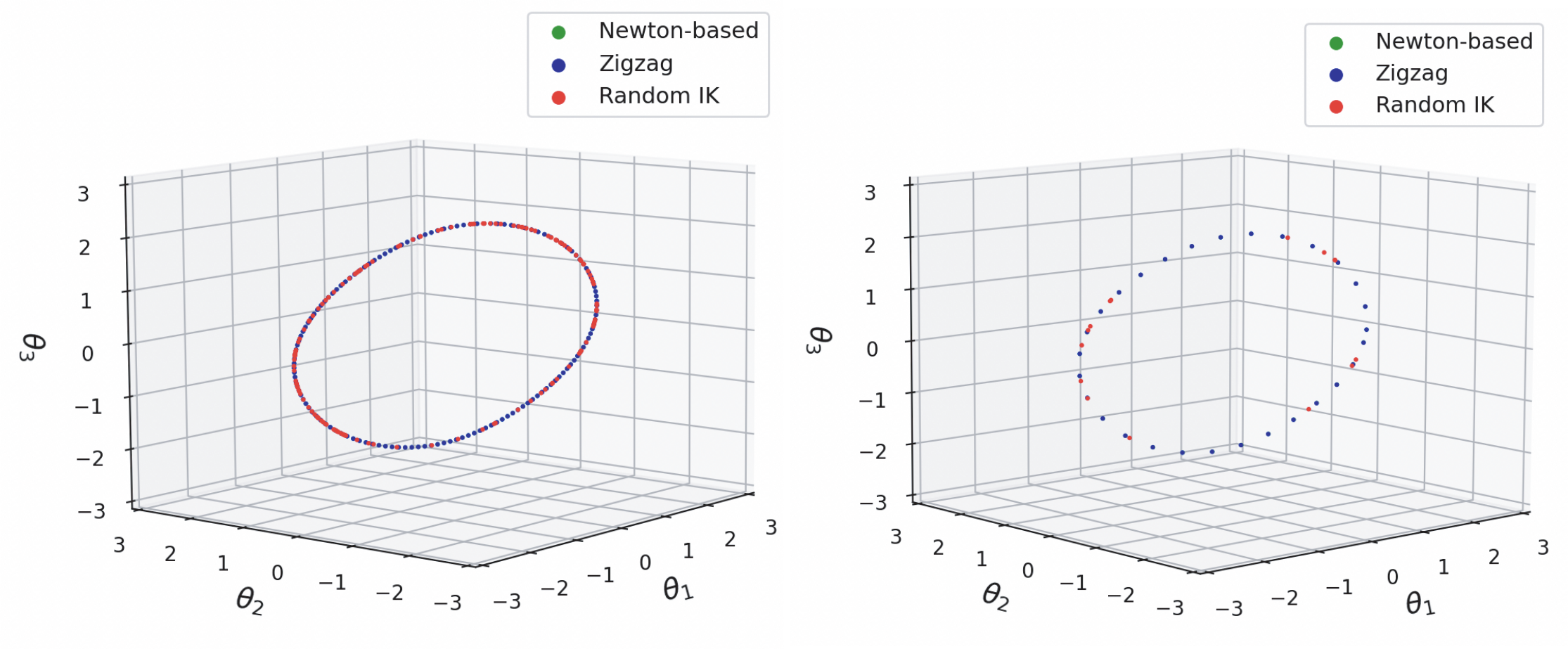}
    \caption{Comparison of sampling results in configuration space.
Left: Jacobian-guided methods ($\beta = 0.1$) vs. Random IK ($n = 150$).
Right: Jacobian-guided methods ($\beta = 1.0$) vs. Random IK ($n = 15$).}
    \label{fig:sampling}
\end{figure}





We compare the Jacobian-guided sampling method proposed in \ref{jacob_guide} with a baseline approach that randomly samples $\mathcal{M}(\bm{p}^{\ast})$ based on inverse kinematics computations.
As a benchmark problem, we consider a 3-DoF planar manipulator and sample points in the neighborhood of the solution manifold corresponding to a target point in a two-dimensional plane.
Since the system has one redundant degree of freedom, the solution set $\mathcal{M}(\bm{p}^{\ast})$ forms a one-dimensional manifold.

In the Jacobian-guided sampling method, the parameter controlling the step size along the tangent direction in \eqref{tangent_update} is set to $\beta = 1.5, 1.0, 0.5, 0.1$.
For the inverse-kinematics-based random sampling method, we choose the number of samples as $n = 15, 30, 150,$ and $300$, which are close to and comparable with those obtained by the Jacobian-guided method.
Fig. \ref{fig:sampling} shows the sampled configurations in the joint-angle space.
The upper figure corresponds to the Random IK method with $n = 150$ and the Jacobian-guided methods (Newton-based and Zigzag) with $\beta = 0.1$.
The lower figure corresponds to the Random IK method with $n = 15$ and the Jacobian-guided methods with $\beta = 1.0$.
As illustrated in Fig. \ref{fig:sampling}, both methods successfully track the one-dimensional solution manifold.

For each method, we report the computation time required for sampling, the occupied volume of the sampled points, and the mean distance from the sampled points to the manifold evaluated using the residual function, as summarized in Table \ref{tab:sampling}.
The coverage volume is approximated by discretizing the configuration space into a uniform grid with spacing $s$ and counting the number of grid points that lie within an $\epsilon$-neighborhood of the sampled points, multiplied by the volume of a single grid cell (in this case $s = 0.05$, $\epsilon = 0.5$).
A larger coverage volume indicates that the solution manifold is covered more broadly.





From the experimental results, we observe that for large values of $\beta$, both the Newton-based and Zigzag methods sample the manifold coarsely, resulting in a small number of samples and reduced computation time.
As $\beta$ decreases, the sampling proceeds with smaller steps along the manifold, increasing the number of sampled points and the computational cost. However, this also leads to a larger covered volume and a smaller average distance from the manifold, indicating improved sampling accuracy.

The Zigzag method is expected to be computationally efficient because it relaxes the manifold projection condition, thereby reducing the cost required for convergence compared to the Newton-based method. As shown in Table \ref{tab:sampling}, under coarse sampling conditions ($\beta = 1.5, 1.0$), the Zigzag method achieves shorter computation times.
In contrast, as $\beta$ becomes smaller, the Newton-based method becomes faster than the Zigzag method.
This behavior can be explained by the number of iterations required for convergence. When $\beta \geq 1.0$, the update in \eqref{tangent_update} induces a large movement along the tangent space, causing the iterate to deviate from the manifold. In this case, the Newton-based method typically requires three to four iterations to project back onto the manifold, whereas the Zigzag method converges within approximately one step.
On the other hand, when $\beta < 1.0$, the tangent-space update remains within the vicinity of the manifold, and even the Newton-based method requires zero or one iteration for projection. Under these conditions, the additional computations in the Zigzag method, such as inner product evaluations at each step, dominate the overall cost, making the Newton-based method faster for smaller values of $\beta$.
Regarding accuracy, the Zigzag method consistently exhibits larger errors measured by the distance to the manifold, which is expected since it does not enforce strict manifold projection at each iteration.


In the random inverse-kinematics (Random IK) sampling method, inverse kinematics is solved $n$ times to find joint configurations $\bm{q}$ satisfying $\bm{p}^{\ast} = \bm{f}(\bm{q})$.
Compared with the Jacobian-guided sampling methods that produce a similar number of samples, the Random IK method requires two to three times more computation time and results in a smaller coverage volume. These results indicate that the proposed Jacobian-guided sampling methods can sample points in the vicinity of the manifold more efficiently.
In Fig. \ref{fig:sampling}, the sampled points obtained by the Newton-based and Zigzag methods overlap visually; however, both methods sample the manifold in a nearly uniform manner.
In contrast, the Random IK method produces noticeable gaps between sampled points, leading to non-uniform coverage of the manifold. This behavior is also reflected in the coverage-volume metric.

Based on the experimental results, we adopt the Newton-based Jacobian-guided sampling method in the following sections.

\begin{table}[t]
\centering
\caption{Computation time for manifold neighborhood sampling and implicit GP construction across two robots and tasks}
\label{tab:null-space_manifold}

\scriptsize
\setlength{\tabcolsep}{1pt}

\begin{tabular}{@{}ccc|ccc@{}}
\multirow{2}{*}{\textbf{Robot}} &
\multirow{2}{*}{\textbf{Task}} & 
\textbf{Redundant} & 
\multirow{2}{*}{\textbf{Samples}} & 
\textbf{Sampling} & 
\textbf{GP Construct}  \\
& & \textbf{DOF}  &  & \textbf{Time [ms]} & \textbf{Time [ms]}\\
\hline
3 DOF arm & position (2D) & 1 & 29 & 2.91 & 0.14 \\
7 DOF arm & position and rotation (6D) & 1 & 34 & 8.43 & 0.14
\end{tabular}

\end{table}

\begin{figure*}[h]
  \centering
  \begin{subfigure}{0.3\linewidth}
    \centering
    \includegraphics[width=\linewidth]{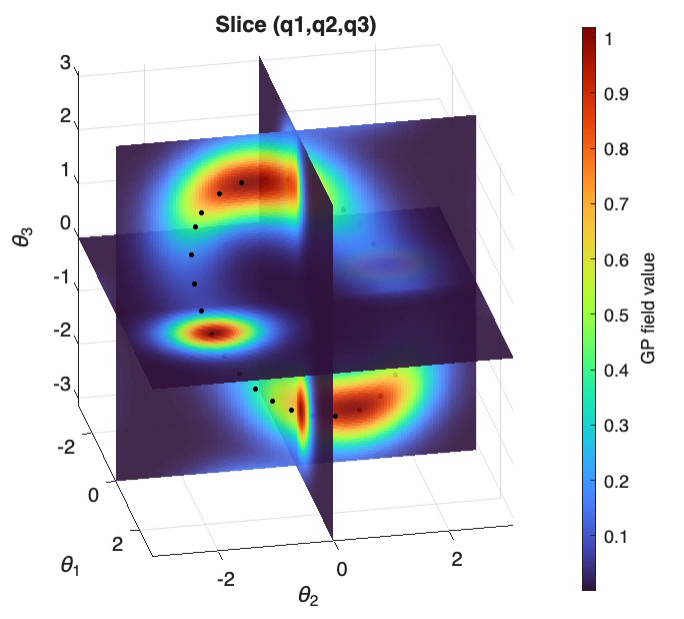}
    \caption{Manifold field}
  \end{subfigure}
  \hfill
  \begin{subfigure}{0.3\linewidth}
    \centering
    \includegraphics[width=\linewidth]{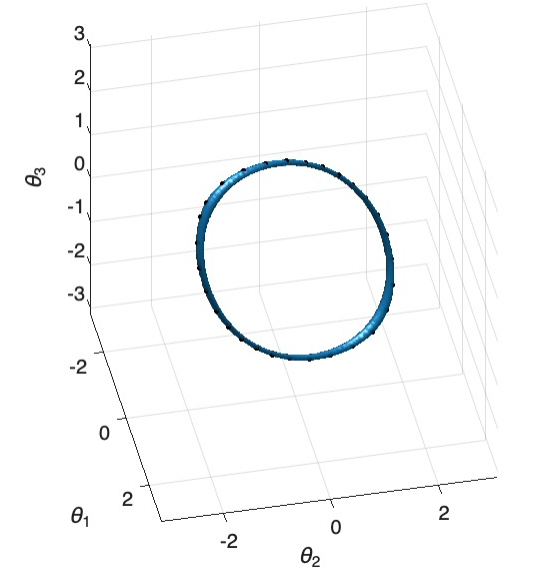}
    \caption{Null-space manifold}
  \end{subfigure}
  \hfill
  \begin{subfigure}{0.3\linewidth}
    \centering
    \includegraphics[width=\linewidth]{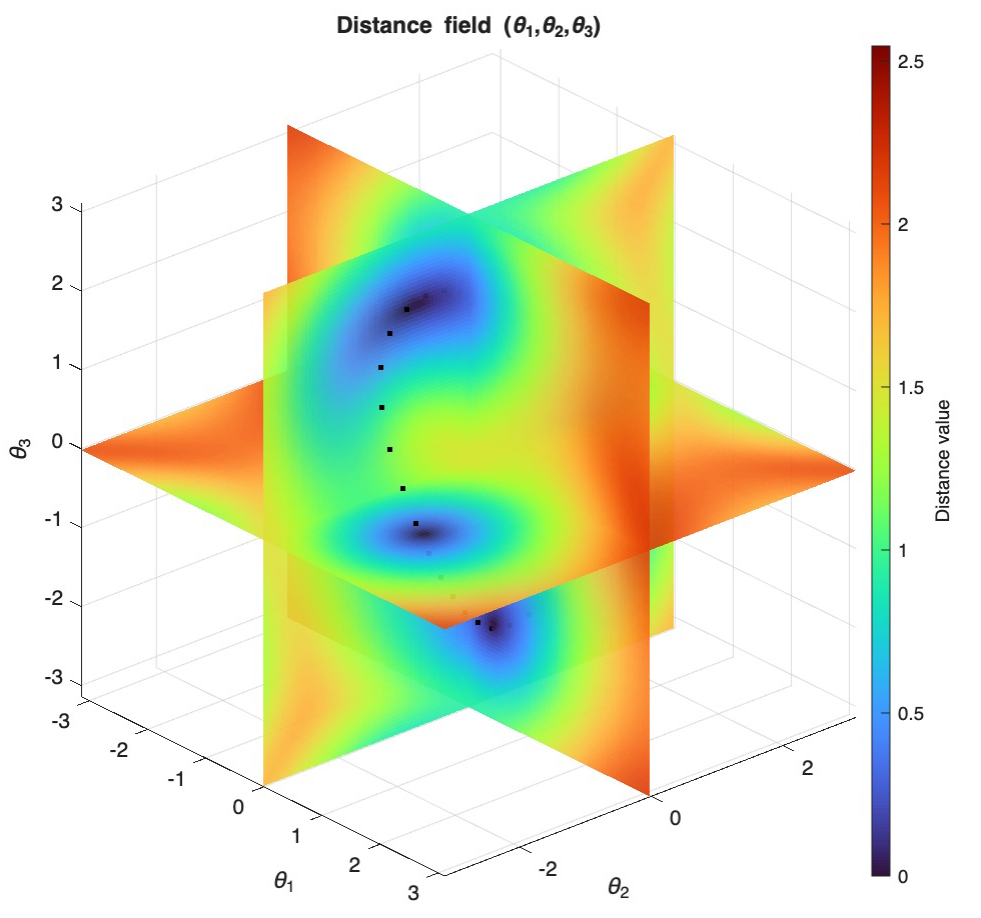}
    \caption{Distance field}
  \end{subfigure}

  \caption{
  Representation of the solution manifold and its induced fields.
  (a) Manifold field. (b) Null-space manifold. (c) Distance field.
  }
  \label{fig:manifold_fields_3Dplanner}
\end{figure*}

\subsection{Generation of Implicit Null-space Manifolds}



In this section, we evaluate whether a representation of the null-space manifold can be obtained from the sampled points generated by the Jacobian-guided sampling method using a Gaussian Process (GP).
Table \ref{tab:null-space_manifold} summarizes the computation times required for Newton-based Jacobian-guided sampling of points in the vicinity of the manifold and for constructing a GP-based manifold representation, for a 3-DoF planar robot arm and a 7-DoF robot arm. The step size for the tangent update in the Newton-based method is set to $\beta = 0.5$.

The computation times required for sampling the manifold neighborhood were 2.91 ms for the 3-DoF planar arm and 8.41 ms for the 7-DoF arm, indicating that the 7-DoF case required approximately 2.89 times more computation time.
In contrast, the time required for GP construction was approximately 0.14 ms for both models. This difference arises because the computational cost of sampling depends on the dimensionality of the configuration space, i.e., the robot’s degrees of freedom, whereas the cost of GP construction mainly depends on the number of sampled points. Since the numbers of sampled points for the two models were similar (29 and 34, respectively), the GP construction times remained nearly identical.



\begin{figure}[t]
    \centering
    \includegraphics[width=\linewidth]{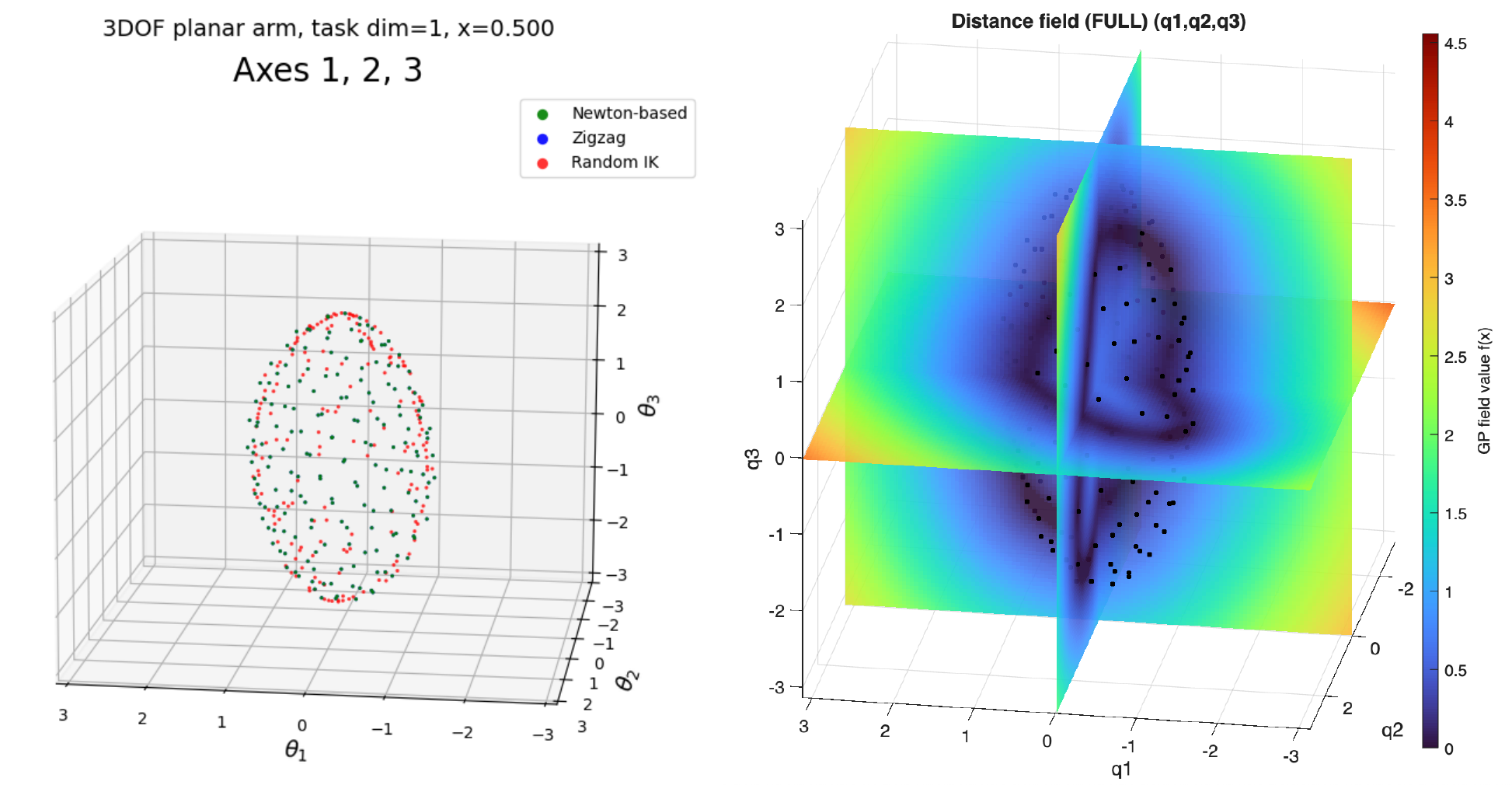}
    \caption{2D null-space manifold and distance field (3D planar robot arm with 1 dof task ($x = 0.05$)).}
    \label{fig:3dof_1dtask}
\end{figure}

We verify whether the constructed null-space manifold is properly represented.
The configuration space is discretized, and inference is performed at each grid point using \eqref{gp_inference}. Since the GP is constructed such that the output value at the sampled points is 1.0, grid points with inference values close to 1 can be regarded as lying on the manifold.
Fig.  \ref{fig:manifold_fields_3Dplanner} (a) shows the inference results obtained by discretizing the configuration space with a resolution of 0.05 for a 3-DoF planar robot arm, where a two-dimensional planar position is specified as the task.
The manifold reconstructed from grid points whose GP inference values exceed 0.995 is shown in Fig. \ref{fig:manifold_fields_3Dplanner} (a). As illustrated in Fig. \ref{fig:manifold_fields_3Dplanner} (b), the one-dimensional null-space manifold is successfully captured.
In addition, Fig.~\ref{fig:manifold_fields_3Dplanner} (c) visualizes the distance field from the manifold computed based on \eqref{distance_form_manifold}.
\added{
To further assess the accuracy of this field, we compare the inferred distances with ground-truth distances obtained via dense sampling for the 1D null-space manifold of the 3-DoF planar arm corresponding to Fig.~\ref{fig:manifold_fields_3Dplanner} (c). The root-mean-square error (RMSE) is $7.67 \times 10^{-3}$, indicating that the proposed method provides an accurate local approximation of the distance field.
}
\added{
The null-space manifold can also be constructed in the case of a 2D redundant manifold, as illustrated in Fig.~\ref{fig:3dof_1dtask}, for a 3-link manipulator.
}

\begin{figure}
    \centering
    \includegraphics[width=0.7\linewidth]{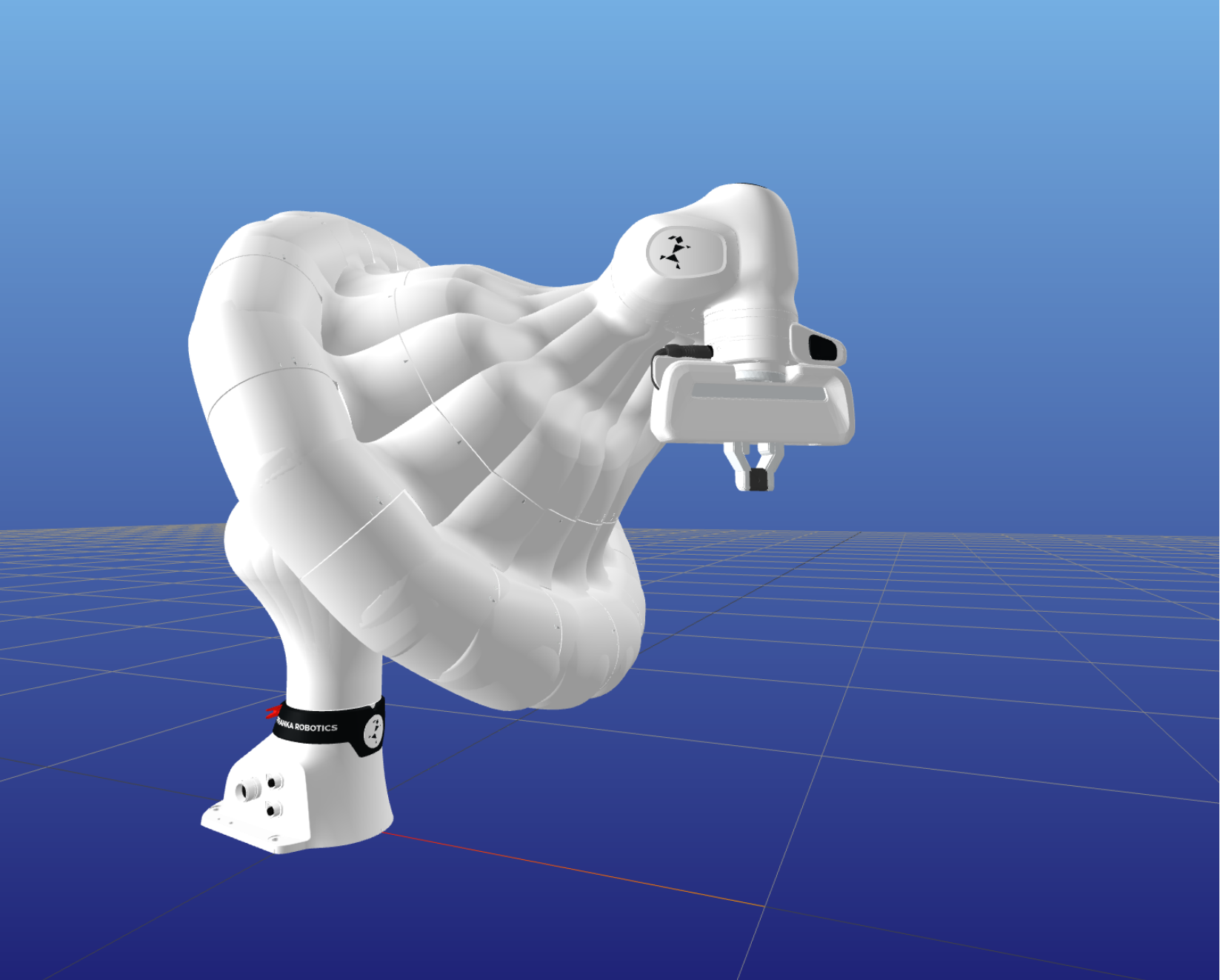}
    \caption{Get poses from null-space manifold.}
    \label{fig:solution manifold}
\end{figure}

Fig. \ref{fig:mani_traj} visualizes the inferred manifold, distance field, and GP values for a 7-DoF robot arm. Since the configuration space is seven-dimensional, three joint axes are selected for visualization, while the remaining dimensions are marginalized by taking the maximum value.
Using the constructed null-space manifold, it is also possible to compute solutions that satisfy the end-effector task, as illustrated in Fig. \ref{fig:solution manifold}.


\subsection{Global Proximity Queries using Implicit Null-space Manifold}



\begin{figure}[t]
  \centering

  \begin{subfigure}{\linewidth}
    \centering
    \includegraphics[width=0.9\linewidth]{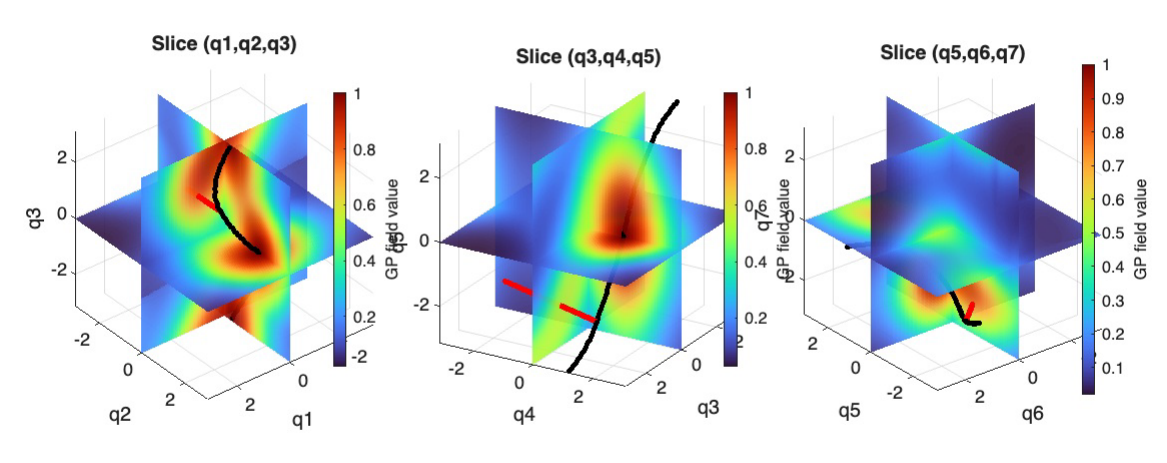}
    \caption{Manifold field}
  \end{subfigure}

  \vspace{0.5em}

  \begin{subfigure}{\linewidth}
    \centering
    \includegraphics[width=0.9\linewidth]{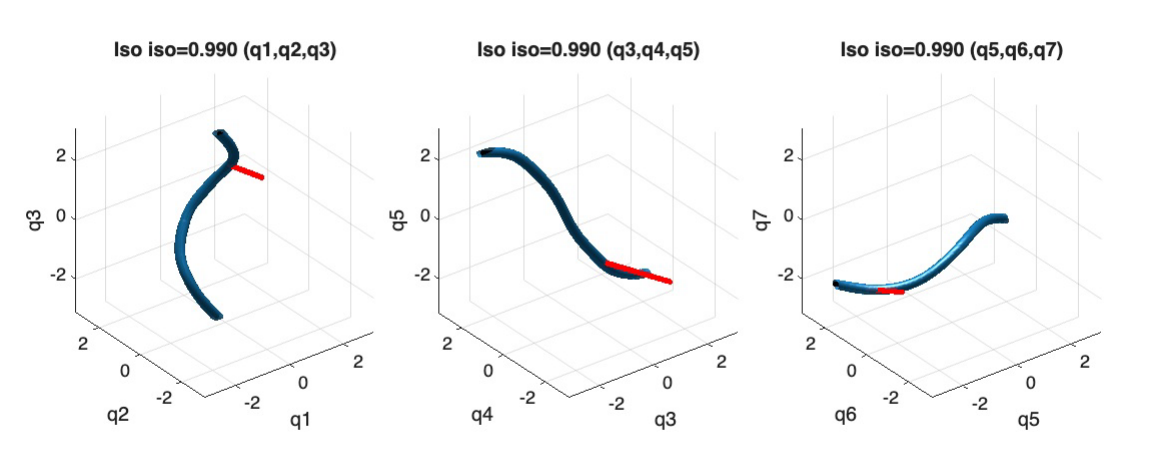}
    \caption{Null-space manifold}
  \end{subfigure}

  \vspace{0.5em}

  \begin{subfigure}{\linewidth}
    \centering
    \includegraphics[width=0.9\linewidth]{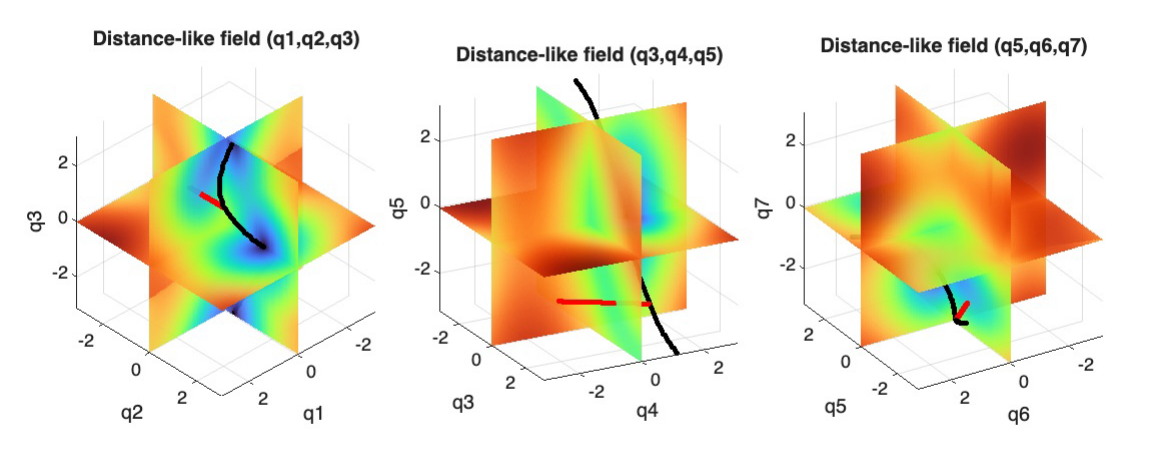}
    \caption{Distance field}
  \end{subfigure}

  \vspace{0.5em}

  \begin{subfigure}{\linewidth}
    \centering
    \includegraphics[width=0.9\linewidth]{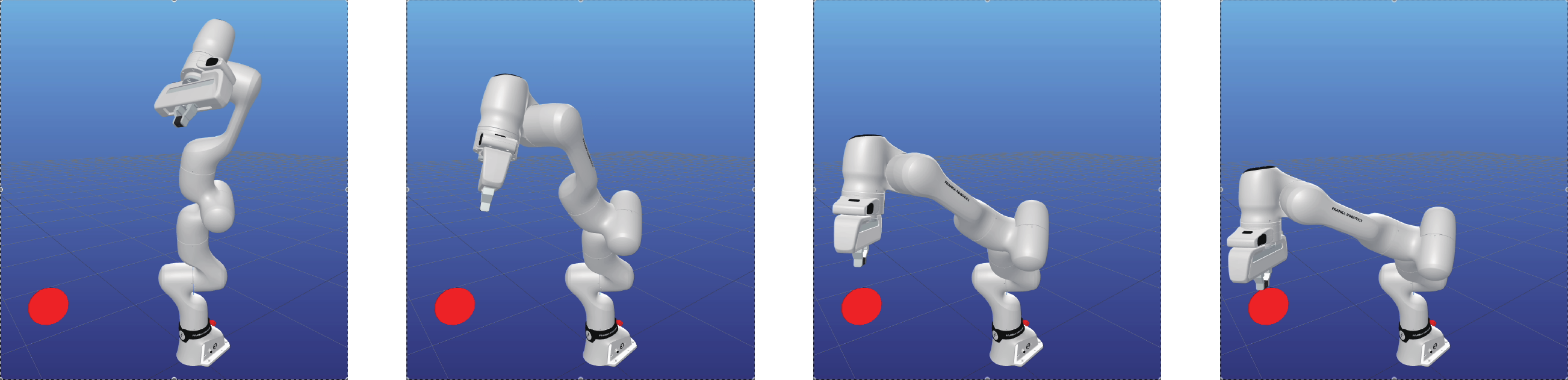}
    \caption{The robot executing planned motion. The induced distance field guides the robot from a distant configuration to the solution manifold.}
  \end{subfigure}

  \caption{Motion generation using the induced distance field. (a)-(c) The visualized distance-like fields and generated trajectories (red lines) in the configuration space. (d) The robot executing the planned motion to reach the target (red dot).}
  \label{fig:mani_traj}
\end{figure}

By using \eqref{distance_form_manifold}, the distance from null-space the manifold can be computed, which makes it possible to obtain the shortest path from an arbitrary pose to the manifold.
Moreover, the direction toward the manifold is given by \eqref{map_gradient}, and moving along this direction by the distance computed from \eqref{distance_form_manifold} allows the configuration to reach the manifold approximately in a single step.
In Figs. \ref{fig:mani_traj} (a), (b), and (c), the red lines indicate trajectories in the configuration space, while in Fig. \ref{fig:mani_traj} (d), the red point represents the target, together with snapshots of the motion trajectory of an actual 7-DoF robot.
Thus, the proposed method enables efficient computation of a path to the manifold.
In practice, this projection step can be computed efficiently, requiring approximately 0.04 ms per evaluation including \eqref{map_gradient} and \eqref{distance_form_manifold}.

\subsection{Reconstruction of Spatially Distributed Tasks in the Configuration Space}

So far, we have focused on tasks defined at a single point in the Euclidean space.
This formulation can be extended to spatially distributed tasks by interpreting \(\bm{p}\) as a field defined over the Euclidean space and reconstructing it in the configuration space within the proposed framework.

We consider two types of spatially distributed tasks: a one-dimensional task defined along a line and a two-dimensional task defined over a rectangular region.
The line task corresponds to finding candidate joint-space trajectories that satisfy a given end-effector trajectory, while the rectangular task represents motions over a planar surface, such as a table-wiping motion.

\begin{figure}[t]
  \centering
  \begin{subfigure}{0.48\linewidth}
    \centering
    \includegraphics[width=\linewidth]{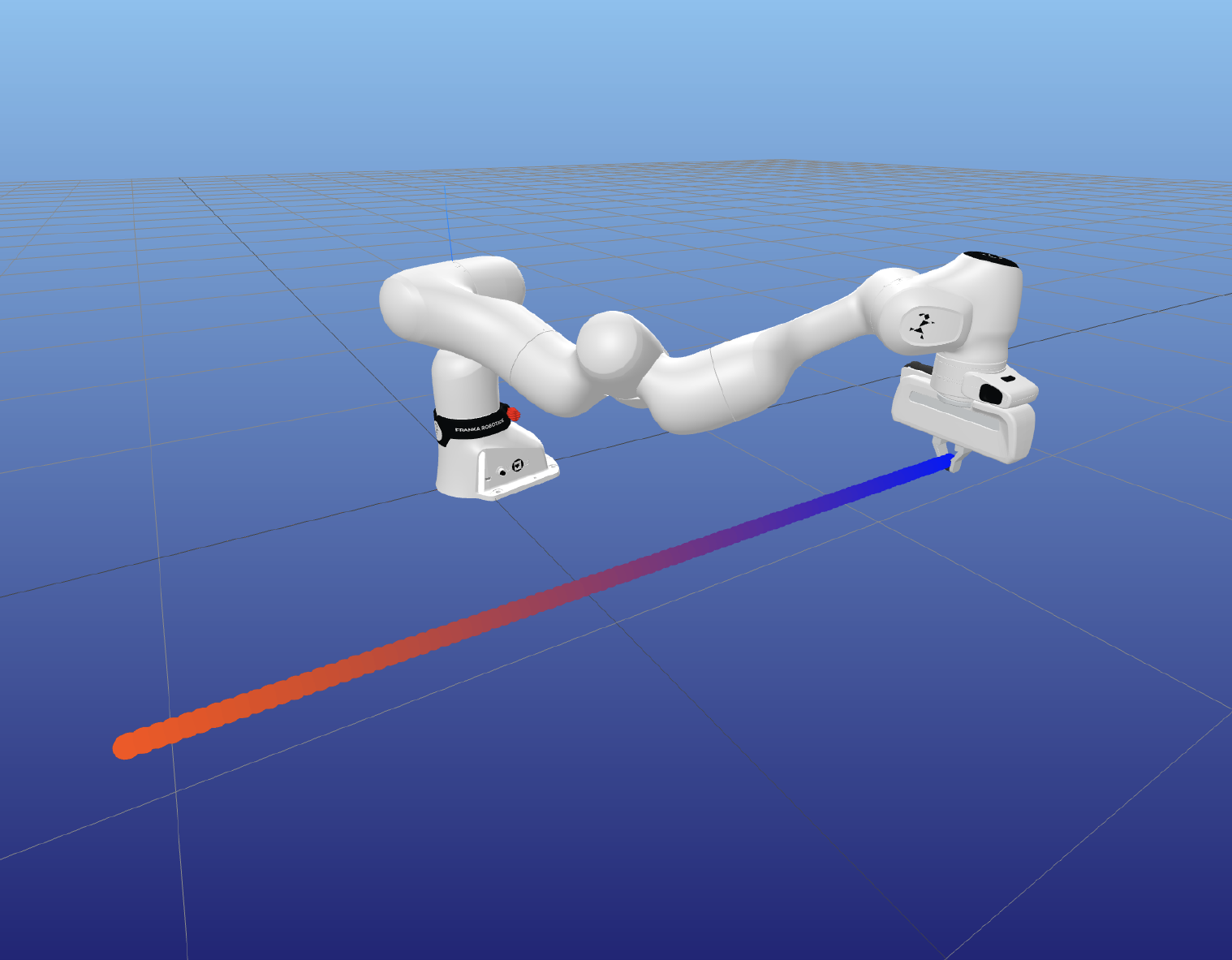}
    \caption{1D line task}
  \end{subfigure}
  \hfill
  \begin{subfigure}{0.48\linewidth}
    \centering
    \includegraphics[width=\linewidth]{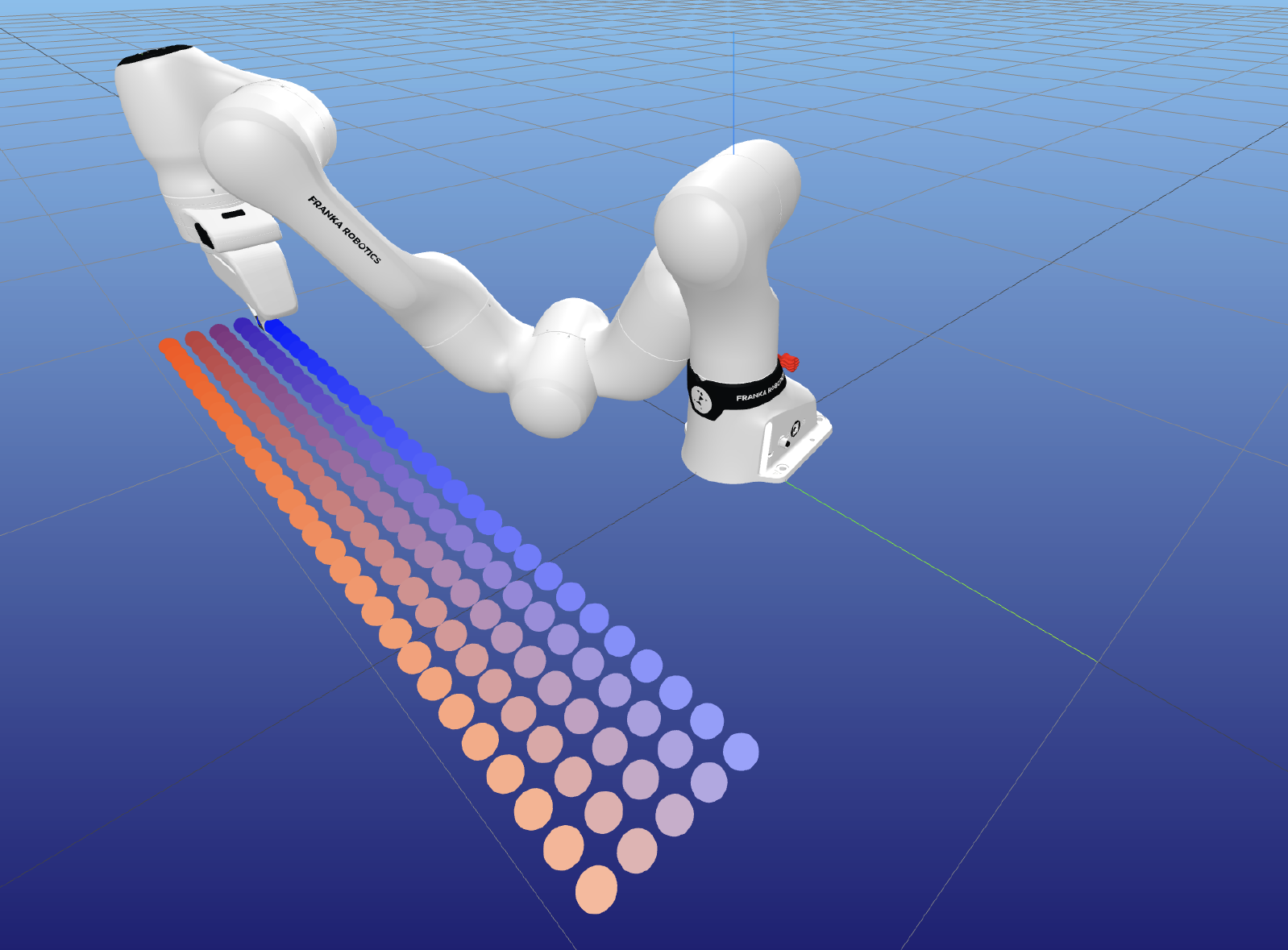}
    \caption{2D planar task}
  \end{subfigure}
  \caption{
Task definitions with continuous variation.
(a) A line task where the end-effector follows a line while maintaining orientation.
(b) A planar task over a rectangular region.
  }
  \label{fig:task_definition}
\end{figure}

\begin{figure}[t]
  \centering
  \begin{subfigure}{\linewidth}
    \centering
    \includegraphics[width=\linewidth]{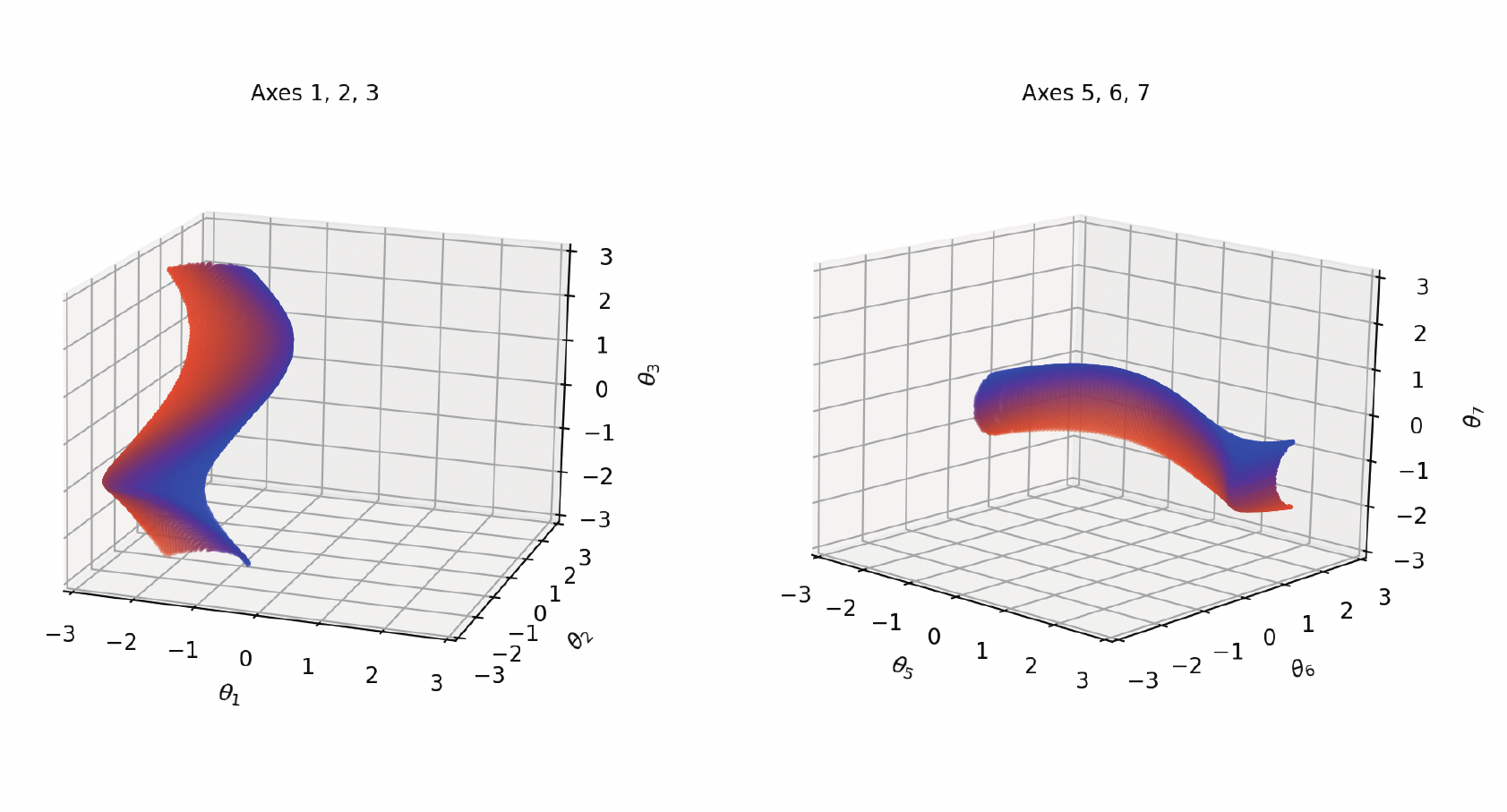}
    \caption{1D task: null-space manifold (colored by task-space position)}
  \end{subfigure}
  \vspace{0.5em}
  \begin{subfigure}{\linewidth}
    \centering
    \includegraphics[width=\linewidth]{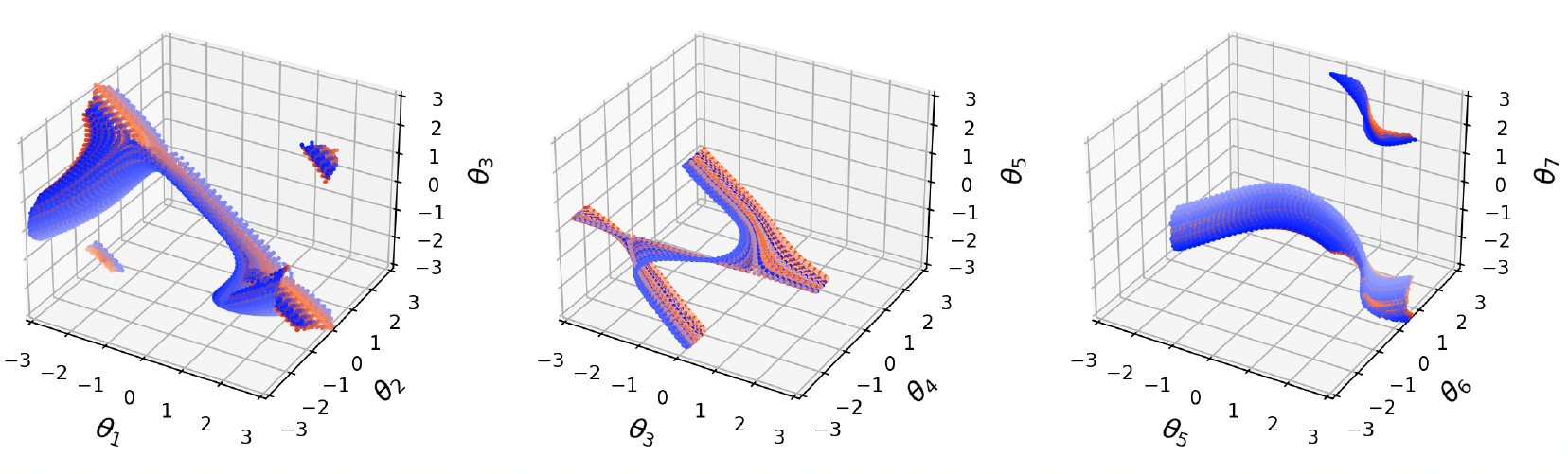}
    \caption{2D task: corresponding 3D null-space manifold}
  \end{subfigure}
  \caption{
Reconstructed manifolds in configuration space.
  }
  \label{fig:manifold_results}
\end{figure}

For the line task, the trajectory is defined as \(y = 0.6,\; -0.5 \leq x \leq 0.5,\; z = 0.3\), and the end-effector of a 7-DoF Franka robot is constrained to maintain a vertically downward orientation (Fig.~\ref{fig:task_definition} (a)).
The null-space manifolds corresponding to the discretized task trajectory (30 points) are visualized simultaneously in Fig.~\ref{fig:manifold_results} (b), where the color indicates the position along the task trajectory.
As a result, a two-dimensional manifold is formed by combining the
one-dimensional task structure with the single redundant degree of freedom at each task point.

Figure~\ref{fig:manifold_results} (b) illustrates the null-space manifold associated with a two-dimensional rectangular task, in which the end-effector maintains a vertically downward orientation while moving over the region
\(0.4 \leq x \leq 0.5,\; -0.4 \leq y \leq 0.4,\; z = 0.3\)
(Fig.~\ref{fig:task_definition} (b)).
The corresponding joint configurations form a three-dimensional manifold, as shown in Fig.~\ref{fig:manifold_results} (b), resulting from the combination of the two-dimensional task structure and one redundant degree of freedom.
The apparent overlap of differently colored points is due to the projection of the seven-dimensional configuration-space manifold onto a three-dimensional space for visualization.
Since the color encodes a field defined on the manifold, this representation suggests that, in the future, it may be possible to represent not only the distance field from the manifold in the configuration space but also fields defined directly on the manifold.

\begin{table}[t]
\centering
\caption{Quantitative Evaluation of Implicit Manifold Representation on 7-DOF Franka Manipulator}
\label{tab:franka_eval}
\begin{tabular}{l|ccccc}
\textbf{Task Type} & \textbf{Samples} & \textbf{\shortstack{Sampling\\Time [s]}} & 
\textbf{\shortstack{GP Construct\\Time [s]}}  & 
\textbf{\shortstack{RMSE\\Error [m]}} \\
\hline
Line Task & 3532 & 0.67 & 0.52 &
$2.56 \times 10^{-5}$ \\
Rectangle Task  & 8738 & 1.59 & 6.41 & 
$1.96 \times 10^{-5}$ \\
\end{tabular}
\end{table}

\section{Discussion}
The proposed framework builds upon classical results on self-motion manifolds, which establish that solution sets induced by redundancy form smooth manifolds in the configuration space \cite{burdick1989inverse}.
While these studies primarily focus on the analytical characterization of manifold properties, our work addresses a complementary and practical question: how to construct a reusable representation of such manifolds that can support repeated queries efficiently.

In conventional inverse kinematics, exploration of the configuration space is performed locally and repeatedly for each query in order to obtain a feasible solution.
In contrast, the proposed approach performs a task-specific exploration of the configuration space in advance and encodes the resulting manifold structure into an implicit representation.
This representation allows subsequent solution queries—such as distance evaluation, gradient computation, and projection onto the solution manifold—to be carried out efficiently, thereby amortizing the cost of exploration across multiple queries.

\added{
From this viewpoint, it is also instructive to compare the proposed approach with existing methods for exploring or representing solution manifolds.
In contrast to ODE-based methods~\cite{guri2025ode}, which trace the manifold along a single tangent direction and are designed for one-dimensional null-space manifolds, the proposed approach incorporates an explicit projection step onto the manifold.
While ODE-based methods rely on fixed step-size integration (e.g., 0.05 in~\cite{guri2025ode}) to maintain constraint accuracy, our approach corrects deviations via projection and remains stable across a wider range of step sizes (e.g., 1.0 to 0.1).
Moreover, extending ODE-based approaches to higher-dimensional null spaces requires handling multiple tangent directions, whereas the proposed framework naturally accommodates such cases.
The method further constructs a reusable implicit representation via sampling and GP-based modeling, enabling efficient exploration and fast proximity queries.
}

Here, reusability refers to the ability to retain the result of this precomputation as a single implicit field and to reuse it across multiple functionalities for a fixed task definition.
Once constructed, the representation supports proximity queries, gradient-based updates, and solution generation without re-solving inverse kinematics from scratch.
In our implementation, such query operations are executed in the order of tens of microseconds, enabling fast online access to solution-space information.

\replaced{
From this perspective, it is instructive to compare the proposed framework with learning-based implicit representations, such as configuration-space distance field methods~\cite{li2024configuration} and related neural-network-based models.
These approaches construct a scalar field over the configuration space that encodes distances to constraint sets, and use gradient-based projection for inverse kinematics and planning.
While both the proposed method and these approaches employ implicit representations, they differ in objective and construction methodology.
Learning-based distance field methods focus on approximating a distance function through offline training, whereas the proposed method directly models the structure of the solution manifold, providing structured access to the solution set for proximity queries, gradient-based updates, and sampling.
Moreover, such learning-based approaches typically rely on computationally expensive offline training procedures, requiring large datasets and training times ranging from hours to days.
In contrast, the proposed method constructs the representation directly from local geometric information obtained via sampling, without requiring iterative training on large datasets.
As a result, the construction time is on the order of milliseconds, substantially reducing the time-to-deployment, particularly in scenarios where task definitions or robot configurations change frequently.
}{
From this perspective, it is also instructive to compare the proposed framework with neural-network-based approaches to inverse or implicit mappings.
Neural methods similarly perform exploration of the configuration space in advance through offline training; however, this exploration typically requires large datasets and computationally expensive training procedures, often spanning hours or days.
In contrast, the proposed method constructs the implicit representation directly from local geometric information, requiring no iterative training on massive datasets.
As a result, the construction time of the representation is on the order of milliseconds, substantially reducing the time-to-deployment, particularly in scenarios where task definitions or robot configurations change frequently.
}

The GPIS formulation adopted in this work plays a key role in enabling this efficient precomputation.
GPIS allows meaningful manifold representations to be constructed from relatively small numbers of samples, which is particularly advantageous when sampling on solution manifolds is costly.
Moreover, GPIS explicitly models uncertainty through predictive variance, providing confidence-aware information about the represented solution space.
The resulting implicit function naturally induces a continuous distance field over the configuration space, from which gradients and projections can be computed directly, supporting efficient query-time operations.

At the same time, the present method is subject to practical limitations that primarily arise from the current strategy for exploring solution manifolds in the configuration space.
Since the proposed framework relies on sampling-based exploration to construct the implicit representation, its scalability is inherently constrained by the curse of dimensionality, and efficient exploration becomes increasingly challenging as the system dimension grows.

\added{
In addition, multiple disconnected solution components may arise in general.
The proposed method explores a single connected component from a given seed using tangent-based traversal and projection.
Additional components can be covered by reinitializing from other feasible configurations, for example obtained via inverse kinematics or random sampling.
}
\added{
Topological transitions may occur near singular configurations, where the tangent space becomes ill-defined, affecting traversal, while projection still enforces approximate constraint satisfaction.
In our 3-DoF planar robot example with a 2D task, the resulting null-space manifold forms a circular structure, and a singular configuration lies at the center of this circle, illustrating such topological behavior.
}
\added{
Joint limits can be incorporated by restricting the configuration domain, while self-collisions are not considered in the current formulation and remain future work.
}

In contrast, the GPIS representation itself does not constitute a fundamental limitation of the framework.
The main computational challenge associated with GPIS lies in the growth of inference cost as the number of samples increases, which is a well-known issue in GP models.
This limitation can be addressed through established techniques such as sparse GP \added{methods based on structured kernel interpolation~\cite{eriksson2018dski}} without altering the underlying representation-centric formulation. 

Finally, while the current formulation targets kinematic task mappings, the underlying viewpoint—precomputing and reusing configuration-space structure through an implicit representation—is not restricted to them.
Extending the framework to incorporate additional constraints, such as obstacle avoidance or interaction conditions, as well as improving scalability through sparse or approximate probabilistic models, constitutes an important direction for future work.
\section{Conclusion}
In this paper, we proposed a representation-centric framework for
explicitly modeling solution manifolds induced by redundancy in
robotic systems. By combining Jacobian-guided sampling with GPIS, the proposed method constructs a
probabilistic implicit representation whose zero-level set
corresponds to the solution manifold in the configuration space.

The main contribution of this work is a shift in emphasis from
conventional point-wise solution computation, such as inverse
kinematics, to reusable spatial representations of solution sets.
The constructed implicit representation is defined over the
configuration space and naturally induces a continuous,
distance field, which facilitates proximity queries,
task-aware sampling, and reuse across related tasks.

While the proposed framework demonstrates promising capabilities,
it also presents several limitations and directions for future
work. In particular, the computational cost of GP
inference poses scalability challenges for high-dimensional systems
and large sample sets. Future research will investigate scalable
probabilistic models and sparse representations. Moreover,
extending the framework to incorporate more complex task
definitions, such as obstacle avoidance and interaction
constraints, remains an important direction toward broader
applicability.

\section*{Acknowledgments}
This work was supported by JSPS KAKENHI Grant Numbers JP22H05002, JP25K21310.
The authors would like to thank Prof. Wael Suleiman for the valuable discussions and insightful comments.

\def\bib_dir{bib}
\bibliographystyle{plainnat}
\bibliography{bib/rss2026}

\end{document}